\documentclass[twoside,11pt]{article}
\usepackage{pgm}
\usepackage{url}
\usepackage{subcaption}
\usepackage{pgfplots}
\usepackage{algorithm}
\usepackage{algorithmic}

\newcommand{\argmax}[2]{\ensuremath{\textrm{argmax}_{#1} #2}}

\newcommand{\pr}{\ensuremath{\mathrm{Pr}}}

\newcommand{\bn}{\ensuremath{\mathcal{B}}}

\newcommand{\mprob}[1] {\pr(#1)}
\newcommand{\cprob}[2] {\pr(#1\;|\;#2)}

\newcommand{\np}{\ensuremath{\mathsf{NP}}}

\newcommand{\nppp}{\ensuremath{\mathsf{NP^{\mathsf{PP}}}}}

\ShortHeadings{Solving MAP with domain knowledge}{Kwisthout}

\begin{document}

\title{Speeding up approximate MAP by applying domain knowledge about relevant variables}

\author{
  \Name{Johan Kwisthout} \Email{johan.kwisthout@donders.ru.nl}\\
  \addr Radboud University, Donders Institute for Brain, Cognition, and Behaviour\\
  \\
  \Name{Andrew Schroeder} \Email{andrew.schroeder@ru.nl}\\
  \addr Radboud University
}

\maketitle

\begin{abstract}
The MAP problem in Bayesian networks is notoriously intractable, even when approximated. In an earlier paper we introduced the Most Frugal Explanation heuristic approach to solving MAP, by partitioning the set of intermediate variables (neither observed nor part of the MAP variables) into a set of relevant variables, which are marginalized out, and irrelevant variables, which will be assigned a sampled value from their domain. In this study we explore whether knowledge about which variables are relevant for a particular query (i.e., domain knowledge) speeds up computation sufficiently to beat both exact MAP as well as approximate MAP while giving reasonably accurate results. Our results are inconclusive, but also show that this probably depends on the specifics of the MAP query, most prominently the number of MAP variables.  
\end{abstract}

\section{Introduction}

In a Bayesian network, the Maximum A Posteriori (MAP) problem is the computational problem of inferring the most probable explanation given evidence, i.e., the mode of a posterior distribution. In a decision support system, where the underlying statistical model is a Bayesian network \citep[e.g.,][]{Dey05,Geenen06,Kuang17,Liu18}, the MAP problem typically establishes the diagnosis or advice that is best supported by the available evidence and as such is a crucial component of such systems. Computing the MAP problem is computationally very demanding in larger networks; the problem is \nppp-hard \citep{Park04} and remains \np-hard under various structural constraints \citep{Campos20,Kwisthout11} as well as under a variety of approximation approaches \citep{Kwisthout15}. This unfavourable complexity not only hinders practical application; it also implies that even the state-of-the-art approximation algorithm ({\sc Annealed MAP}; \cite{Yuan04}) will have difficulty on at least {\em some} problem instances.

To partially overcome this challenge, in earlier work we proposed a {\em heuristic} approach to MAP, based on the observation that in many real-world inference queries, only a small subset of the variables really contributes to the decision. In the {\sc Most Frugal Explanation} heuristic \citep{Kwisthout15a} the set of intermediate variables in the network (those variables that are neither to be explained nor contain observations) is partitioned, based on background information about their role in the inference process, into relevant and irrelevant variables. The relevant variables are marginalized over, the irrelevant variables are sampled from. By this method, and under some assumptions about the probability distribution, a reasonable approximation of the MAP explanation can be offered with hopefully less resources. We showed that the quality of the explanation (that is, its deviation from the ground truth MAP explanation in terms of structural distance, rank, and probability) strongly depends on the accuracy of the partitioning in relevant and irrelevant variables, as well as on the probability landscape.\\

In this follow-up paper we are interested in the question to what extent {\em domain knowledge} about which variables are and are not relevant in a particular query can help with speeding up the MAP computation while maintaining reasonable accuracy. To that end, we compared the {\sc Most Frugal Explanation} heuristic with {\sc Exact MAP} and {\sc Annealed MAP} (ANN) on a number of benchmark networks, measuring the amount of time needed for computation and the deviation from the actual MAP explanation. In general, for an arbitrary computation only a small proportion of the intermediate variables will be relevant \citep{Druzdzel94}. Knowledge of the causal structure modelled by the Bayesian network, previous experience, or clinical evidence can help to assess this relevance. In addition, in the absence of an authoritative source of information, one can {\em compute}, on the fly, whether a particular variable is instrumental for a particular inference question, or use a pre-computed assessment in a lookup table. This computation aims to assess the likeliness (expressed as {\em intrinsic relevance}, see Section \ref{Preliminaries}) that a variable, were it observed to one of the possible values in its domain, would change the outcome of the MAP problem. In this paper we investigate both aspects: Is knowledge about relevance helpful for establishing MAP explanations in situations where we 1) have access to a `lookup table' and 2) approximately compute this on the fly as part of the heuristic? The algorithm that will be used for the first situation where a lookup table is provided will be referred to as MFE+ throughout the rest of the paper, while the algorithm that will be used to compute the relevance of variables on the fly will simply be called MFE. Obviously, the second method takes more time as it is integrated in the heuristic. More specifically, our research questions and working hypotheses are the following:

\begin{itemize}
\item \textbf{RQ 1:} If we have access to pre-computed knowledge about which intermediate variables are irrelevant (have a low likeliness of changing the outcome of the MAP problem), could this speed up approximate inference, with comparable accuracy, compared to the state-of-the-art approximation algorithm?
\item \textbf{Hypothesis 1:} Yes, on large networks with a huge number of intermediate variables, and with many of them being irrelevant, it will. In either other case probably not.
\item \textbf{RQ 2:} If we assess, by sampling, during the execution of the algorithm which intermediate variables are relevant, could this speed up approximate inference, with comparable accuracy, compared to the state-of-the-art approximation algorithm?
\item \textbf{Hypothesis 2:} It might, under the same conditions as in RQ 1, if the assessment of the relevance of the variables is computationally cheap. Otherwise, probably not.
\end{itemize}

The latter condition is probably necessary. Computing intrinsic relevance is an \np-hard problem \citep{Kwisthout15a}. In this study we investigate how costly this computation is in a number of benchmark cases, to see whether this may be a viable option at all.\\

In addition to these initial research questions, it was hypothesized that it may be possible to speed up the MFE+ algorithm by using ANN to internally approximate the MAP value rather than using the exact MAP algorithm. This would take the advantages of MFE+ (utilization of pre-computed variable relevance) and the advantages of the ANN approximation algorithm and ideally produce an algorithm that was faster than the vanilla ANN and MFE+ algorithms, albeit with additional approximation error. In a sense the idea was to use an approximation algorithm inside another approximation algorithm to test if the there was an advantageous trade-off between reduced computational time and increased error. For brevity this new algorithm will be referred to as MFE+A and the intent of it is captured in the third research question:
\begin{itemize}
    \item \textbf{RQ 3:} Given the MFE+ algorithm (where pre-computed relevance values are provided) is it possible to speed up this algorithm by using an annealed MAP approximation instead of the exact MAP solution? If it is faster, is it worth using considering the additional error that is incurred?

    \item \textbf{Hypothesis 3:} The algorithm may be faster, but the additional error incurred is likely to render such a method infeasible in practice.
\end{itemize}

The remainder of this paper is structured as follows. In Section \ref{Preliminaries} we formally introduce the MAP problem in Bayesian networks, describe {\sc Annealed MAP} and the {\sc Most Frugal Explanation} heuristic, and share our notational conventions. In Section \ref{Methods} we explicate the experimental setup of this study, describe the characteristics of the benchmark networks, and motivate some of the choices made in the study. The results are given and discussed in Section \ref{Results}. We conclude the paper in Section \ref{Conclusion}.

\section{Preliminaries}
\label{Preliminaries}

A Bayesian network $\bn$ is a probabilistic graphical model that describes a set of stochastic variables, a joint probability distribution over these variables, and the conditional independences that hold in this distribution \citep{Pearl88}. $\bn$ includes a directed acyclic graph $\mathbf{G}_{\mathcal{B}} = (\mathbf{V}, \mathbf{A})$, modelling the variables and conditional independences in the network, and a set of parameter probabilities $\pr$ in the form of conditional probability tables (CPTs), capturing the strengths of the stochastic relationships between the variables. The network efficiently factorizes a joint probability distribution $\mprob{\mathbf{V}} = \prod_{i=1}^n \cprob{V_i}{\pi(V_i)}$ over its variables, where $\pi(V_i)$ denotes the parents of $V_i$ in $\mathbf{G}_{\mathcal{B}}$. As notational convention we will use upper case letters to denote individual nodes in the network, upper case bold letters to denote sets of nodes, lower case letters to denote value assignments to nodes, and lower case bold letters to denote joint value assignments to sets of nodes. The set of values for a particular variable (and by extension, set of variables) $V$ is denoted as $\Omega(V)$.

Given a partitioning of the variables in the network into explanation variables $\mathbf{H}$, evidence variables $\mathbf{E}$, and intermediate variables $\mathbf{I}$, the MAP problem is the computational problem to establish the best explanation $\mathbf{h}$ to $\mathbf{H}$ given an observation $\mathbf{e}$ to $\mathbf{E}$. More formally, the MAP problem is the problem to find $\argmax{\mathbf{h}}{ \mprob{\mathbf{H}=\mathbf{h},\mathbf{E}=\mathbf{e}}}$. The MAP problem is \nppp-hard and thus highly intractable \citep{Park04}. It stays \np-hard even in trees with cardinality $3$ \citep{Campos20}, yet enjoys a fixed parameter tractable algorithm for several parameter sets, e.g., when the tree-width of the moralization of the network, the cardinality of the variables, and the size of $\mathbf{H}$ is bounded.

Among the approximation algorithms are for example {\sc Anytime Approximate MAP} \citep{Maua12}, {\sc Annealed MAP} \citep{Yuan04}, and {\sc P-Loc} \citep{Park01}. In this paper we focus on comparison with {\sc Annealed MAP}, a local search algorithm which can, with good accuracy, approximate relatively large benchmark networks. Approximate MAP, however, is \np-hard as well if no additional constraints are imposed on the input \citep{Kwisthout11,Kwisthout15a,Park04}. 

In \cite{Kwisthout15a} we introduced a novel approach towards MAP where we exploit a generic property of probability distributions: often-times, when making an inference only a small subset of all variables are actually instrumental in the computation \citep{Druzdzel94}. Assuming a partition of the intermediate variables $\mathbf{I}$ in a set of relevant variables $\mathbf{I^+}$ and irrelevant variables $\mathbf{I^-}$, the MFE heuristic approaches MAP by sampling over the irrelevant variables and marginalizing only over the relevant variables. This heuristic, because of this sampling, cannot deal well with deterministic variables as this might lead to conflicting evidence. The algorithm for MFE is given below, taken from \cite[p. 64]{Kwisthout15a}.

In order to formally assess the relevance of intermediate variables, the concept {\em Intrinsic relevance} was defined in \cite{Kwisthout15a} as the fraction of joint value assignments $\mathbf{i}$ in $\Omega(\mathbf{I} \setminus \{I\})$ for which $\argmax{\mathbf{h}}{\mprob{\mathbf{h},\mathbf{e},\mathbf{i},i}}$ is not identical for all $i \in \Omega(I)$.

\begin{algorithm}[H]
\label{algo_MFE}
\caption{Compute the Most Frugal Explanation}
Sampled-MFE$(\bn, \mathbf{H}, \mathbf{I^+}, \mathbf{I^-}, \mathbf{e}, N)$
\begin{algorithmic}[1]
\FOR{$n = 1$ to $N$}
\STATE Choose $\mathbf{i} \in \mathbf{I^-}$ at random
\STATE Determine $\mathbf{h} = \argmax{\mathbf{h}}{\mprob{\mathbf{H} = \mathbf{h}, \mathbf{i}, \mathbf{e}}}$
\STATE Collate the joint value assignments $\mathbf{h}$
\ENDFOR
\STATE Decide upon the joint value assignment $\mathbf{h}_{\mathrm{maj}}$ that was picked most often
\RETURN $\mathbf{h}_{\mathrm{maj}}$
\end{algorithmic}
\end{algorithm}

\section{Methods}
\label{Methods}

In order to evaluate the {\sc Most Frugal Explanation} (hereafter MFE) we studied the performance of this heuristic in comparison with Exact MAP via the {\sc Junction Tree} algorithm (hereafter MAP) and the {\sc Annealed MAP} algorithm (hereafter ANN) on several benchmark Bayesian networks ({\sc Alarm}, {\sc Andes}, {\sc Barley}, and {\sc Hailfinder}). We implemented\footnote{\url{https://gitlab.socsci.ru.nl/j.kwisthout/most-frugal-explanations}.} both MFE and ANN in C++ using the LibDAI library \citep{Mooij10} and compared running time and accuracy in terms of the number of variables in the explanation set that differed from the actual MAP explanation. We ran additional tests to check for differences between different accuracy measures, to assess the impact of the hypothesis space, and to compare theoretical and run-time results. 

\subsection{Experimental setup}

We ran our algorithms on an HP Compaq Elite 8300 CMT desktop computer, with Intel Core i7-3770 CPU running at 3.40 GHz and 16 GB of memory, running Debian GNU/Linux 10. We partitioned the variables of the Bayesian network into hypothesis variables, evidence variables, and intermediate variables as described per benchmark network below. For each network we randomly assigned ten joint value assignments to the evidence nodes and simulated each MAP query five times to average out perturbations in running time due to external factors (such as OS activity). We computed the Hamming distance between the MAP explanation and the heuristic approaches (hereafter denoted as `error') and averaged running time and error over the $5 \times 10 = 50$ simulations. Data was locally stored and processed; raw data, scripts, and processed data are available\footnote{\url{https://doi.org/10.34973/7p10-m012}} at the Donders Data Repository for colleagues to inspect and reuse.

Per benchmark network, we ported the .bif files from the BNLearn repository\footnote{\url{https://www.bnlearn.com/bnrepository/}} to factor graphs using an in-house tool\footnote{{\sc Bif2Fg}, \url{https://gitlab.socsci.ru.nl/j.kwisthout/most-frugal-explanations}; note that this tool assumes a specific ordering of the CPT entries to work.} since the LibDAI library requires factor graphs as input. As MFE cannot deal properly with deterministic variables, we manually adjusted these variables to have values very close to $1$ and $0$; each $0$ entry in a CPT was replaced with $0.000000001$ and the $1$ entry matched such that the distribution adds up to $1$. We use the orginal factor graphs for MAP and ANN and the adjusted factor graphs for MFE computations\footnote{Note that none of the algorithms exploits determinism, and the values used have a negligible impact.}.

For the MFE heuristic we simulated two variants: one where the partitioning into relevant and irrelevant variables was given (by pre-computation) as to simulate background knowledge, and one where the partitioning was part of the heuristic approach. In the first case, as pre-computation we approximated the intrinsic relevance by $1000$ samples or, in case that was computationally infeasible, by $100$ or even $10$, as indicated per benchmark network below; the threshold for inclusion in the set of relevant variables was set to $0.1$. In the second case, the approximation was part of the heuristic; we sampled thrice and deemed a variable as relevant if the intrinsic relevance was non-zero\footnote{Given the actual distribution, any threshold between $0$ and $0.9$ would have led to identical results in these experiments. The threshold for inclusion, however, is a tunable parameter in MFE in general.}. In both variants, the algorithm marginalized out the relevant variables and assigned a random value to the irrelevant variables.

The LibDAI library has no functionality to compute MAP efficiently; only inference (using the junction tree algorithm \citep{Lauritzen88}) and MPE (that is, when there are no intermediate variables). As a workaround we computed MAP by computing the joint distribution over the MAP variables and then searching for the maximum value. Obviously this workaround approach does not utilize the independences between the hypothesis variables, leading to an inefficient implementation that hugely effects the results; see Section \ref{Running_time_vs_theoretical_analysis} for a further analysis. To mitigate this, we decided to limit the size of the hypothesis set; rather than the $20$ hypothesis variables in \cite{Yuan04} we selected five variables in {\sc Andes} and {\sc Hailfinder} and four in {\sc Barley}. For {\sc Hailfinder} we explored the effect of including more variables (specifically seven and ten) in the hypothesis set\footnote{One reviewer suggested that the number of candidate hypotheses, rather than the size of the hypothesis set (i.e., the number of variables), is the crucial variable. That is a correct observation; however, given minor differences between the hypothesis variables we actually considered, the size of the hypothesis set is the most dominant factor determining the number of candidate hypothesis.}. Note that approximate MAP is intractable, even when a singleton binary hypothesis variable is used \citep{Kwisthout15}.

\subsubsection{Alarm}

The {\sc Alarm} network \citep{Beinlich89} consists of $37$ discrete random variables which have a natural partitioning into hypothesis variables (eight diagnostic variables), evidence variables (sixteen observable findings) and intermediate variables (the remaining thirteen variables). Variables can take on two, three, or at most four values; there are $46$ arcs and the in-degree is at most four. In our simulations we used the natural partitioning as indicated above. Pre-computation of the intrinsic relevance of the thirteen intermediate variables was done using $1000$ samples each; pre-computation time varied between $11$ and $78$ seconds per variable.

\subsubsection{Barley}

The {\sc Barley} network \citep{Kristensen02} has $48$ nodes with $84$ arcs, with in-degree at most four. However, the cardinality of the variables is significantly larger, with one node having no less than $67$ states. This renders exact MAP infeasible on the {\sc Barley} network. In line with \cite{Yuan04} we interpreted the ten root variables as hypothesis variables and the eight leaf variables as evidence variables, which reasonably matches the layout of this network \cite[p.206]{Kristensen02}. Due to the impossibility to compute either MAP or MFE over the total hypothesis set due to the library constraints, we selected the first four variables as our hypothesis set. From piloting it became pretty clear that pre-computation of intrinsic relevance using $1000$ samples was infeasible, as a single sample already took about $90$ seconds due to the intractability of MAP computation. We approximated intrinsic relevance using $10$ samples per variable.

\subsubsection{Andes}

{\sc Andes} \citep{Conati97} has $223$ binary nodes with $338$ arcs, and in-degree at most six; the CPTs are to a large extent rolled out from canonical models (leaky noisy or/and models). There are $89$ root nodes and $22$ leaf nodes; for these simulations, the first five root variables make up the hypothesis set. Furthermore, as {\sc Andes} has several deterministic variables, for the MFE computation we adjusted these variables as indicated above. Due to the large number of intermediate variables we opted for $100$ samples per variable to approximate intrinsic relevance.

\subsubsection{Hailfinder}

Finally, {\sc Hailfinder} \citep{Abramson96} has $56$ nodes with $66$ arcs with maximum in-degree four. This network has $17$ root nodes and $13$ leaf nodes; the cardinality ranges from two to eleven. Again the size of the hypothesis set renders exact MAP intractable in practice. To experiment with the effect of the size of the hypothesis set on the behaviour of the different algorithms we ran simulations using the first $10$, $7$, and $5$ root variables as hypothesis set. Also for {\sc Hailfinder} we adjusted the deterministic variables as indicated above. In contrast to MAP inference, computing the intrinsic relevance using $1000$ samples was feasible, taking from $10$ to $60$ seconds.

\section{Results}
\label{Results}

\begin{figure}[h!] 
\centering
\begin{subfigure}{0.50\textwidth}
\pgfplotsset{width=7.2cm}
\begin{tikzpicture}
\begin{semilogyaxis} [
	y tick label style={/pgf/number format/fixed zerofill},
	symbolic x coords={MAP,ANN,MFE,MFE+},
	xtick distance=1,
	ylabel=Running time (secs),
	enlargelimits=0.15,
	legend style={at={(0.5,-0.15)},	anchor=north,legend columns=-1},
	ybar,
	ymin=0.001,
	log origin=infty,
	log ticks with fixed point,
	bar width=7pt,
]
\addplot coordinates {(MAP,0.0135) (ANN,0.1536) (MFE,0.1546) (MFE+,0.0133)};
\addplot coordinates {(MAP,1.473) (ANN,120.9530) (MFE,2420.1568) (MFE+,1.4669)};
\addplot coordinates {(MAP,0.1556) (ANN,6.3393) (MFE,188.1614) (MFE+,0.1511)};
\addplot coordinates {(MAP,0.0051) (ANN,0.2891) (MFE,1.5000) (MFE+,0.0061)};
\legend{{\sc Alarm},{\sc Barley},{\sc Andes},{\sc Hail5}}
\end{semilogyaxis}
\end{tikzpicture}
\end{subfigure}
\hspace{0.5cm}
\begin{subfigure}{0.45\textwidth}
\pgfplotsset{width=6.8cm}
\begin{tikzpicture}
\begin{axis} [
	y tick label style={/pgf/number format/fixed},
	symbolic x coords={MAP,ANN,MFE,MFE+},
	xtick distance=1,
	ylabel=Errors,
	enlargelimits=0.15,
	ybar,
	bar width=7pt,
]
\addplot coordinates {(MAP,0) (ANN,0.08) (MFE,0.42) (MFE+,0.70)};
\addplot coordinates {(MAP,0) (ANN,0.46) (MFE,1.38) (MFE+,1.66)};
\addplot coordinates {(MAP,0) (ANN,0.04) (MFE,0.26) (MFE+,0.56)};
\addplot coordinates {(MAP,0) (ANN,0.70) (MFE,1.42) (MFE+,1.84)};
\end{axis}
\end{tikzpicture}
\end{subfigure}
\caption{Graphical depiction of the main results for running time (RT) and error (Err) for the four approaches on four benchmark networks. In the left panel from left to right the average running time in seconds for exact MAP, Annealed MAP, MFE with sampled relevance, and MFE with pre-computed relevance (note the log scale); in the right panel the average errors.}
\label{main_results_fig}
\rule{\columnwidth}{0.3mm}
\vspace{1mm}
\end{figure}
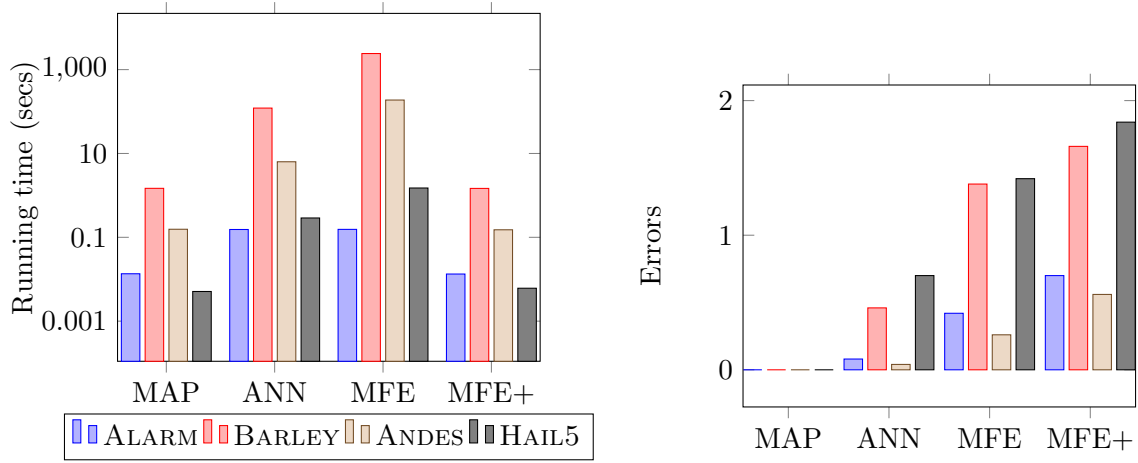

\begin{figure}[h!] 
\centering
\begin{subfigure}{0.47\textwidth}
\pgfplotsset{width=7.0cm}
\begin{tikzpicture}
\begin{semilogyaxis} [
	y tick label style={/pgf/number format/fixed zerofill},
	xtick distance=1,
	symbolic x coords={MAP,ANN,MFE,MFE+},
	ylabel=Running time (secs),
	enlargelimits=0.15,
	legend style={at={(0.5,-0.15)},	anchor=north,legend columns=-1},
	ybar,
	ymin=0.001,
	log origin=infty,
	log ticks with fixed point,
	bar width=7pt,
]
\addplot coordinates {(MAP,0.0051) (ANN,0.2891) (MFE,1.5000) (MFE+,0.0061)};
\addplot coordinates {(MAP,0.0290) (ANN,0.5799) (MFE,1.4581) (MFE+,0.0290)};
\addplot coordinates {(MAP,20.3924) (ANN,0.6144) (MFE,21.5747) (MFE+,20.2491)};
\legend{H5,H7,H10}
\end{semilogyaxis}
\end{tikzpicture}
\end{subfigure}
\hspace{0.5cm}
\begin{subfigure}{0.47\textwidth}
\pgfplotsset{width=7.0cm}
\begin{tikzpicture}
\begin{axis} [
	y tick label style={/pgf/number format/fixed},
	symbolic x coords={MAP,ANN,MFE,MFE+},
	xtick distance=1,
	ylabel=Errors,
	enlargelimits=0.15,
	ybar,
	bar width=7pt,
]
\addplot coordinates {(MAP,0) (ANN,0.70) (MFE,1.42) (MFE+,1.84)};
\addplot coordinates {(MAP,0) (ANN,0.90) (MFE,2.56) (MFE+,2.44)};
\addplot coordinates {(MAP,0) (ANN,1.30) (MFE,3.00) (MFE+,3.02)};
\end{axis}
\end{tikzpicture}
\end{subfigure}
\caption{Results for the {\sc Hailfinder} network with $5$, $7$, and $10$ hypothesis nodes. In the left panel from left to right the average running time in seconds for exact MAP, Annealed MAP, MFE with sampled relevance, and MFE with pre-computed relevance (note the log-scale); in the right panel the average errors. Note that for $10$ hypothesis nodes the inefficiency of the MAP computation dominates the running times.}
\label{hailfinder_result_fig}
\rule{\columnwidth}{0.3mm}
\vspace{1mm}
\end{figure}
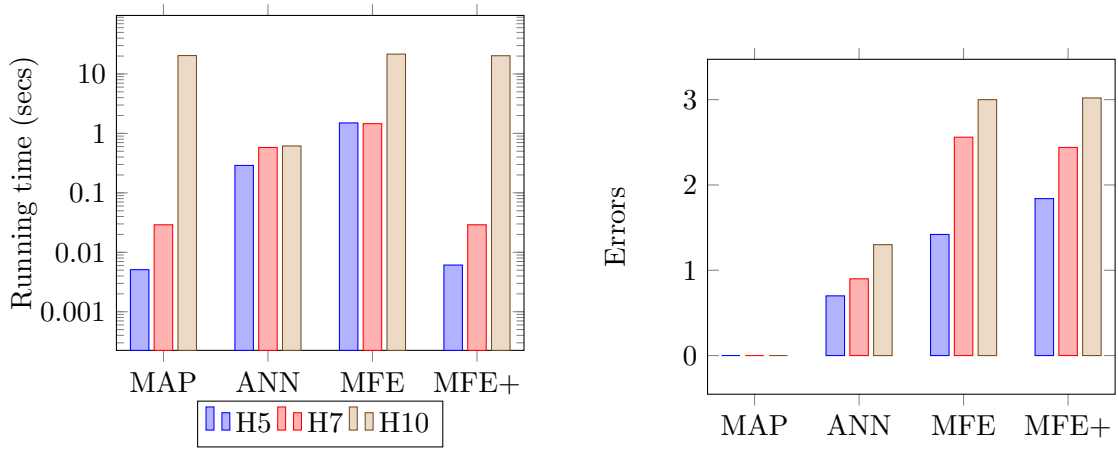

This result section is structured as follows. We start with presenting the main results for \textbf{RQ 1} and \textbf{RQ 2}, namely running time and accuracy of MAP, ANN, MFE with sampling, and MFE with pre-computation (MFE+), for {\sc Alarm}, {\sc Barley}, {\sc Andes}, and {\sc Hailfinder} (with five hypothesis variables) in Table \ref{table_main_results} and graphically in Figure \ref{main_results_fig}. We then further investigate the effect on hypothesis set size for {\sc Hailfinder} (Figure \ref{hailfinder_result_fig}) and present a preliminary interpretation of these results. In Sub-section \ref{Other_error_measures} we investigate, using the {\sc Alarm} network, whether the Hamming distance results sufficiently generalize to other accuracy measures for MAP. In Sub-section \ref{Size_of_relevant_variables_set} we look at the number of variables that are actually considered to be relevant (based on the pre-computation) per network, and in Sub-section \ref{Running_time_vs_theoretical_analysis} we use this information to compare actual running time with some theoretical analyses of the number of elementary operations needed. We discuss the results for \textbf{RQ 2} and MFE+A in Sub-section \ref{results_for_research_q3} before discussing the overall simulations in Sub-section \ref{Discussion}.

\begin{table}[ht!]
\begin{tabular}{|c|cc|cc|cc|cc|}
\hline
Network & \multicolumn{2}{c|}{\sc Alarm} & \multicolumn{2}{c|}{\sc Barley} & \multicolumn{2}{c|}{\sc Andes} & \multicolumn{2}{c|}{\sc Hailfinder (5)}\\
 & RT & Err & RT & Err & RT & Err & RT & Err\\    
\hline
MAP  & 0.0135 & -    &    1.4973 & -    &   0.1556 & -    & 0.0051 & -    \\
ANN  & 0.1536 & 0.08 &  120.9530 & 0.46 &   6.3393 & 0.04 & 0.2891 & 0.70 \\
MFE  & 0.1546 & 0.42 & 2420.1568 & 1.38 & 188.1614 & 0.26 & 1.5000 & 1.42 \\
MFE+ & 0.0133 & 0.70 &    1.4669 & 1.66 &   0.1511 & 0.56 & 0.0061 & 1.84 \\
\hline
\end{tabular}
\caption{Summary of the main results for running time (RT) and error (Err) for the four approaches on four benchmark networks.}
\label{table_main_results}
\end{table}

The results are difficult to interpret. For this hypothesis size, MAP is quicker than ANN and MFE and as quick as MFE+! It seems that the initial costs and the estimation of effect on local changes for ANN are too large to be competitive for MAP for small hypothesis sizes; for MFE the sampling is way too costly, which may be due to an unreasonable large MAP computation time due to the limitations of the library. We further explored this in Section \ref{Running_time_vs_theoretical_analysis}. For MFE+, we see that the limited amount of marginalisation operations has little effect on the running time.

When comparing errors, we note that, to our surprise, in MFE the sampling over $3$ samples led to less errors than sampling over $1000$ samples in the pre-compute stage. The only reasons we can think of that explain this would be that by either the sampling result is incorrect, or that the heuristic sometimes (even with perfect knowledge) leads to a different MFE explanation than the MAP explanation, as also observed in \cite{Kwisthout15a}, such that the lower number of samples incidentally leads to an explanation closer to MAP.

As a follow-up simulation we explored the effect of a larger hypothesis space. For the {\sc Hailfinder} network we ran the same simulations, but now also with the first seven and the first ten root variables designated as hypothesis variables. The results are depicted in Figure \ref{hailfinder_result_fig}.

\subsection{Other error measures}
\label{Other_error_measures}

\begin{figure}[h!] 
\centering
\pgfplotsset{width=7.0cm}
\begin{tikzpicture}
\begin{semilogyaxis} [
	symbolic x coords={ANN,MFE,MFE+},
	xtick distance=1,
	enlargelimits=0.15,
	y tick label style={/pgf/number format/fixed},
	ybar,
	bar width=9pt,
	ylabel=Accuracy measurement,
	legend style={at={(0.5,-0.15)},	anchor=north,legend columns=-1},
	log origin=infty,
	log ticks with fixed point
]
\addplot coordinates {(ANN,0.08) (MFE,0.42) (MFE+,0.70)};
\addplot coordinates {(ANN,0.96) (MFE,0.70) (MFE+,0.48)};
\addplot coordinates {(ANN,1.10) (MFE,4.88) (MFE+,8.72)};
\legend{Hamming,Ratio,Rank}
\end{semilogyaxis}
\end{tikzpicture}
\caption{Comparison between Hamming distance, ratio, and rank of the explanations. Note that for distance and rank lower is better, whereas for ratio a value closer to $1$ is better.}
\label{accuracy_comparison_fig}
\rule{\columnwidth}{0.3mm}
\vspace{1mm}
\end{figure}
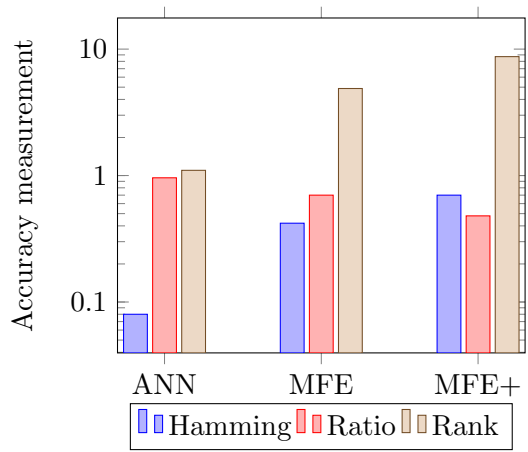

In \cite{Kwisthout15a}, three error measures were used, in line with the three notions of approximate MAP established in \cite{Kwisthout15}; in addition to the Hamming distance between ground truth MAP and the approximate explanation, the {\em ratio} of their probabilities as well as the {\em rank} of the explanation (i.e., the number $k$ such that the approximate explanation is the $k$th most probable explanation) were used. These measures can in principle deviate, hence, we ran a sanity check test to assure that the Hamming distance gives a reasonable impression of the quality of the approximate explanation. The results can be found in Figure \ref{accuracy_comparison_fig}.

ANN scores best on all three measures, then MFE, then MFE+. This suggests that indeed the Hamming distance gives a good indicator of the accuracy of the approximation.

\subsection{Size of relevant variables set}
\label{Size_of_relevant_variables_set}

\begin{figure}[h!] 
\centering
\pgfplotsset{width=8.5cm}
\begin{tikzpicture}
\begin{axis} [
	y tick label style={/pgf/number format/fixed},
	symbolic x coords={Alarm,Andes,Barley,Hail5,Hail7,Hail10},
	ylabel=Relevant variables,
	enlargelimits=0.15,
	ybar,
	bar width=10pt,
]
\addplot coordinates {(Alarm,0.36) (Andes,0.03) (Barley,0.09) (Hail5,0.05) (Hail7,0.08) (Hail10,0.20)};
\end{axis}
\end{tikzpicture}
\caption{Ratio of relevant variables out of all intermediate variables.}
\label{relevant_ratio_fig}
\rule{\columnwidth}{0.3mm}
\vspace{1mm}
\end{figure}
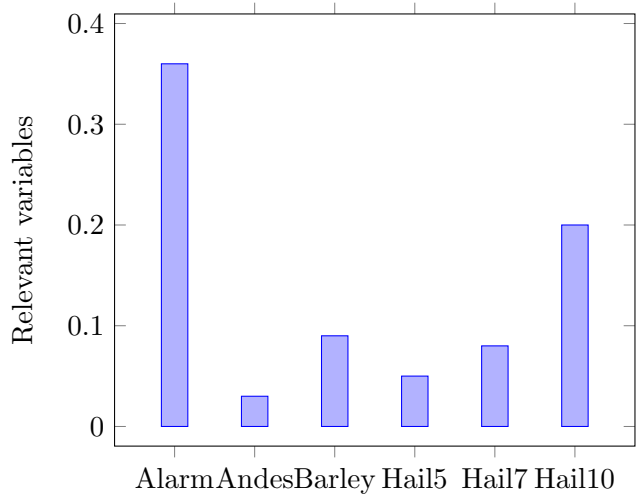

We refer to Figure \ref{relevant_ratio_fig}. In general: With more hypothesis variables, the ratio increases (see the increase in {\sc Hailfinder} but also the large ratio in {\em Alarm}). For smaller sets, the number of relevant variables is really low (see for example {\em Andes}), which ought to lead to a more efficient algorithm; reducing factors with a large set of sampled values potentially greatly reduces the resulting factor size and discards the marginalization computation. However, in reality, it does not, at least not for these simulations. We explore this in the next Sub-section. 

\subsection{Running time vs. theoretical analysis}
\label{Running_time_vs_theoretical_analysis}

For the Alarm network, we see that on average four out of the eleven relevant variables are relevant. We closely examined a particular run to get some insight in the discrepancy between theoretical results and actual runtime. Table \ref{alarm_inspection_run} shows the characteristics of the particular MAP query and MFE+ heuristic.

\begin{table}[h!]
\begin{tabular}{llllll}
variable & value & variable & value & variable & value\\
\hline
HISTORY & FALSE & CVP & HIGH & PCWP & HIGH \\
HRBP & HIGH & HREKG & NORMAL & HRSAT & NORMAL\\
TPR & HIGH & EXPCO2 & HIGH & MINVOL & ZERO\\
FIO2 & LOW & SAO2 & LOW & PAP & NORMAL\\
PRESS & NORMAL & MINVOLSET & LOW & CO & LOW\\
BP & LOW \\
\end{tabular}
\caption{Example evidence setting.}
\label{alarm_inspection_run}
\end{table}

For this query, LVEDVOLUME, STROKEVOLUME, VENTTUBE, and VENTALV were found relevant; we sample over ERRLOWOUTPUT, ERRCAUTER, PVSAT, SHUNT, VENTMACH, VENTLUNG, ARTCO2, CATECHOL, and HR. The biggest factor, once they are reduced with the evidence and sampled intermediate variables, now contains KINKEDTUBE, INTUBATION, and VENTTUBE or $2 \times 3 \times 4 = 24$ entries, rather than four variables with $92$ entries. When comparing MAP to MFE+ we see that they LibDAI library spends $2.1ms$ on a junction tree run and $12.2ms$ computing the marginal distribution in MAP, with $0.8ms$, respectively $8.2ms$ for MFE+. The library does not utilize the fact that, after marginalization, the biggest factor contains two hypothesis variables with 12 entries (rather than all hypothesis variables with $384$ entries), which may explain why we see less effect on the running times than what could be expected in theory.

\subsection{Results for Research Question 3: MFE+A}
\label{results_for_research_q3}

The experimental setup was the same as that described in section 3.1 - the experiment was run on the same computer and each network had a total of 50 simulations (10 different evidence values averaged across 5 runs, per network). The only difference is that the MFE+A algorithm was added and the regular MFE algorithm was removed to save on simulation time.

\begin{figure}[h!] 
\centering
\begin{subfigure}{0.45\textwidth}
\pgfplotsset{width=7.2cm}
\begin{tikzpicture}
\begin{semilogyaxis} [
	y tick label style={/pgf/number format/fixed zerofill},
	symbolic x coords={MAP,ANN,MFE+A,MFE+},
	xtick distance=1,
	ylabel=Running time (secs),
	enlargelimits=0.15,
	legend style={at={(0.5,-0.15)},	anchor=north,legend columns=-1},
	ybar,
	ymin=0.001,
	log origin=infty,
	log ticks with fixed point,
	bar width=7pt,
]

\addplot coordinates {(MAP,0.0145145132) (ANN,0.16107993172) (MFE+A,0.00840941384) (MFE+,0.0097581622)};
\addplot coordinates {(MAP,1.56088961946) (ANN,126.14093631922) (MFE+A,7.32085905058) (MFE+,1.5319950777)};
\addplot coordinates {(MAP,0.159820465933333) (ANN,6.38019308671111) (MFE+A,0.350601094177778) (MFE+,0.155472659333333)};
\addplot coordinates {(MAP,0.0058108099) (ANN,0.66127244612) (MFE+A,0.01383215968) (MFE+,0.00366218108)};
\legend{{\sc Alarm},{\sc Barley},{\sc Andes},{\sc Hail5}}
\end{semilogyaxis}
\end{tikzpicture}
\end{subfigure}
\hspace{0.5cm}
\begin{subfigure}{0.45\textwidth}
\pgfplotsset{width=7.2cm}
\begin{tikzpicture}
\begin{axis} [
	y tick label style={/pgf/number format/fixed},
	symbolic x coords={MAP,ANN,MFE+A,MFE+},
	xtick distance=1,
	ylabel=Errors,
	enlargelimits=0.15,
	ybar,
	bar width=7pt,
]
\addplot coordinates {(MAP,0) (ANN,0.06) (MFE+A,1.18) (MFE+,0.66)};
\addplot coordinates {(MAP,0) (ANN,0.62) (MFE+A,1.72) (MFE+,1.68)};
\addplot coordinates {(MAP,0) (ANN,0) (MFE+A,0.69) (MFE+,0.44)};
\addplot coordinates {(MAP,0) (ANN,1) (MFE+A,3.08) (MFE+,1.58)};
\end{axis}
\end{tikzpicture}
\end{subfigure}
\caption{Graphical depiction of the main results for running time (RT) and hamming error (Err) for the four algorithms on four benchmark networks. In the left panel from left to right the average running time in seconds for exact MAP, Annealed MAP, MFE+A, and MFE with pre-computed relevance (note the log scale); in the right panel the average errors.}
\label{main_results_fig_MFE+A}
\rule{\columnwidth}{0.3mm}
\label{ABAHResults}
\vspace{1mm}
\end{figure}
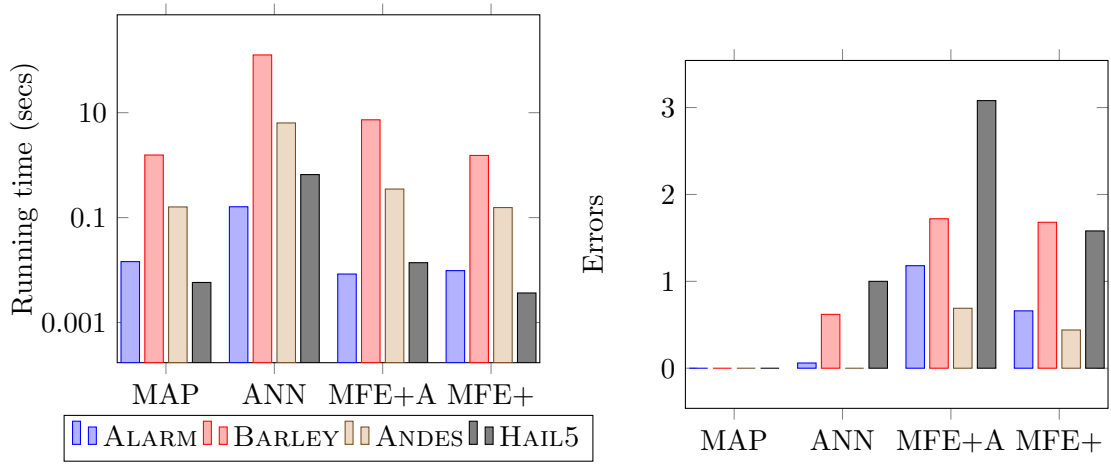
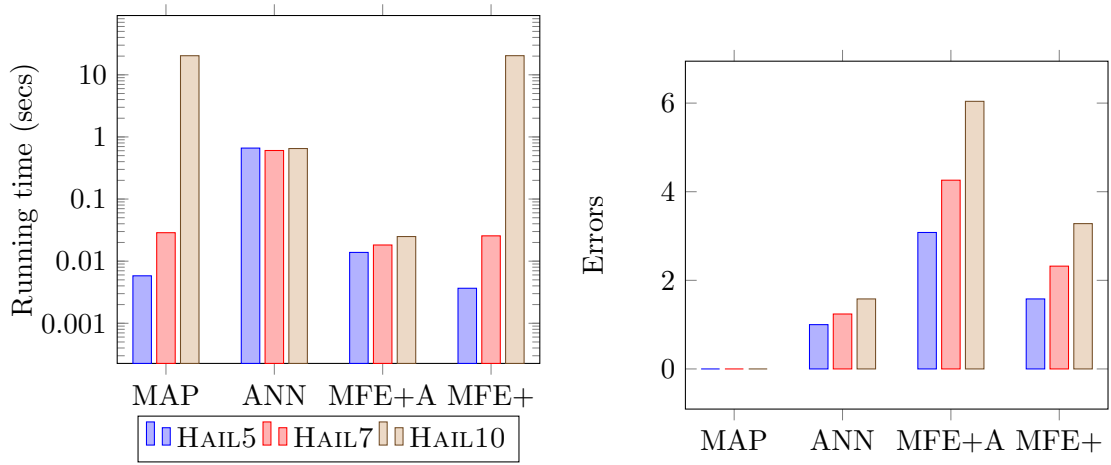
\begin{figure}[h!] 
\centering
\begin{subfigure}{0.45\textwidth}
\pgfplotsset{width=7.2cm}
\begin{tikzpicture}
\begin{semilogyaxis} [
	y tick label style={/pgf/number format/fixed zerofill},
	symbolic x coords={MAP,ANN,MFE+A,MFE+},
	xtick distance=1,
	ylabel=Running time (secs),
	enlargelimits=0.15,
	legend style={at={(0.5,-0.15)},	anchor=north,legend columns=-1},
	ybar,
	ymin=0.001,
	log origin=infty,
	log ticks with fixed point,
	bar width=7pt,
]
\addplot coordinates {(MAP,0.0058108099) (ANN,0.66127244612) (MFE+A,0.01383215968) (MFE+,0.00366218108)};
\addplot coordinates {(MAP,0.0287242298) (ANN,0.6047266944) (MFE+A,0.01820995778) (MFE+,0.025559676)};
\addplot coordinates {(MAP,20.29923141998) (ANN,0.65068806496) (MFE+A,0.02490812672) (MFE+,20.33515500326)};
\legend{{\sc Hail5},{\sc Hail7},{\sc Hail10}}
\end{semilogyaxis}
\end{tikzpicture}
\end{subfigure}
\hspace{0.5cm}
\begin{subfigure}{0.45\textwidth}
\pgfplotsset{width=7.2cm}
\begin{tikzpicture}
\begin{axis} [
	y tick label style={/pgf/number format/fixed},
	symbolic x coords={MAP,ANN,MFE+A,MFE+},
	xtick distance=1,
	ylabel=Errors,
	enlargelimits=0.15,
	ybar,
	bar width=7pt,
]
\addplot coordinates {(MAP,0) (ANN,1) (MFE+A,3.08) (MFE+,1.58)};
\addplot coordinates {(MAP,0) (ANN,1.24) (MFE+A,4.26) (MFE+,2.32)};
\addplot coordinates {(MAP,0) (ANN,1.58) (MFE+A,6.04) (MFE+,3.28)};
\end{axis}
\end{tikzpicture}
\end{subfigure}
\caption{Results for the {\sc Hailfinder} network with $5$, $7$, and $10$ hypothesis nodes. In the left panel from left to right the average running time in seconds for exact MAP, Annealed MAP, MFE+A, and MFE with pre-computed relevance (note the log-scale); in the right panel the average errors.}
\rule{\columnwidth}{0.3mm}
\label{HHHResults}
\vspace{1mm}
\end{figure}

Examining the left sub-figure in Figure \ref{ABAHResults}, it is clear that MFE+A performs better on the {\sc Alarm}, {\sc Barley}, {\sc Andes}, and {\sc Hailfinder5} networks compared with ANN, however it performs slightly worse than MFE+. As expected though the error for MFE+A is higher than either ANN or MFE+ individually when looking at the Hamming error measure. From this figure we can conclude that at least on the initial four networks, MFE+A does not provide any advantage and MFE+ or ANN should be preferred. How does MFE+A perform on the larger {\sc Hailfinder} networks though? The answer is presented in Figure \ref{HHHResults}. In the left sub-figure MFE+A clearly outperforms ANN by several orders of magnitude and also significantly outperforms MFE+ on the largest network - {\sc Hailfinder}10, this is promising! However, despite the running time being quite good, the error measure in the right sub-figure is much higher for MFE+A than either of the two individual algorithms. \\

As expected there is a large improvement in running time with a concomitant rise in the Hamming error, at least for the largest of networks. We have seen in Sub-section \ref{Other_error_measures} that the Hamming distance gives a reasonable impression of the quality of the approximate explanation. Despite this, the authors were curious if MFE+A would perform any better when using the other error metrics and all three error metrics were calculated for each network and are presented in Figure \ref{ErrorMetrics_MFE+A}. Unfortunately it does not appear that MFE+A performs much better even under the new error metrics. Across all six 6 networks, MFE+A undoubtedly has the least accuracy across all three accuracy scores.

\newpage
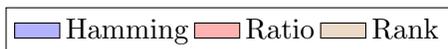
\begin{figure}[H] 
\centering
\begin{subfigure}{0.45\textwidth}
\pgfplotsset{width=7.2cm}
\centering
\begin{tikzpicture}
\begin{semilogyaxis} [
	symbolic x coords={ANN,MFE+A,MFE+},
	xtick distance=1,
	enlargelimits=0.15,
	y tick label style={/pgf/number format/fixed},
	ybar,
	bar width=9pt,
	ylabel=Accuracy measurement,
	legend style={at={(0.5,-0.15)},	anchor=north,legend columns=-1},
	log origin=infty,
	log ticks with fixed point
]
\addplot coordinates {(ANN,0.06) (MFE+A,1.18) (MFE+,0.66)};
\addplot coordinates {(ANN,0.965053283756968) (MFE+A,0.342368402778585) (MFE+,0.57304635541639)};
\addplot coordinates {(ANN,0.12) (MFE+A,20.66) (MFE+,8.38)};
\end{semilogyaxis}
\end{tikzpicture}
\caption{ALARM}
\end{subfigure}
\hspace{0.5cm}
%
\begin{subfigure}{0.45\textwidth}
\pgfplotsset{width=7.2cm}
\begin{tikzpicture}
\begin{semilogyaxis} [
	symbolic x coords={ANN,MFE+A,MFE+},
	xtick distance=1,
	enlargelimits=0.15,
	y tick label style={/pgf/number format/fixed},
	ybar,
	bar width=9pt,
	legend style={at={(0.5,-0.15)},	anchor=north,legend columns=-1},
	log origin=infty,
	log ticks with fixed point
]
\addplot coordinates {(ANN,0.12) (MFE+A,0.74) (MFE+,0.56)};
\addplot coordinates {(ANN,0.882446597488297) (MFE+A,0.467052626982824) (MFE+,0.528135296382728)};
\addplot coordinates {(ANN,0.44) (MFE+A,3.74) (MFE+,2.7)};
\end{semilogyaxis}
\end{tikzpicture}
\caption{ANDES}
\end{subfigure}

\begin{subfigure}{0.45\textwidth}
\pgfplotsset{width=7.2cm}
\centering
\begin{tikzpicture}
\begin{semilogyaxis} [
	symbolic x coords={ANN,MFE+A,MFE+},
	xtick distance=1,
	enlargelimits=0.15,
	y tick label style={/pgf/number format/fixed},
	ybar,
	bar width=9pt,
	ylabel=Accuracy measurement,
	legend style={at={(0.5,-0.15)},	anchor=north,legend columns=-1},
	log origin=infty,
	log ticks with fixed point
]
\addplot coordinates {(ANN,0.62) (MFE+A,1.72) (MFE+,1.68)};
\addplot coordinates {(ANN,0.953675653375436) (MFE+A,0.439663660641988) (MFE+,0.43455022939135)};
\addplot coordinates {(ANN,7.16) (MFE+A,116.88) (MFE+,114.28)};
\end{semilogyaxis}
\end{tikzpicture}
\caption{BARLEY}
\end{subfigure}
\hspace{0.5cm}
%
\begin{subfigure}{0.45\textwidth}
\pgfplotsset{width=7.2cm}
\begin{tikzpicture}
\begin{semilogyaxis} [
	symbolic x coords={ANN,MFE+A,MFE+},
	xtick distance=1,
	enlargelimits=0.15,
	y tick label style={/pgf/number format/fixed},
	ybar,
	bar width=9pt,
	legend style={at={(0.5,-0.15)},	anchor=north,legend columns=-1},
	log origin=infty,
	log ticks with fixed point
]
\addplot coordinates {(ANN,1) (MFE+A,3.08) (MFE+,1.58)};
\addplot coordinates {(ANN,0.974463839620674) (MFE+A,0.225351728803124) (MFE+,0.644020757986495)};
\addplot coordinates {(ANN,2.94) (MFE+A,295.92) (MFE+,29.44)};
\end{semilogyaxis}
\end{tikzpicture}
\caption{HAIL5}
\end{subfigure}

\begin{subfigure}{0.45\textwidth}
\pgfplotsset{width=7.2cm}
\centering
\begin{tikzpicture}
\begin{semilogyaxis} [
	symbolic x coords={ANN,MFE+A,MFE+},
	xtick distance=1,
	enlargelimits=0.15,
	y tick label style={/pgf/number format/fixed},
	ybar,
	bar width=9pt,
	ylabel=Accuracy measurement,
	legend style={at={(0.5,-0.15)},	anchor=north,legend columns=-1},
	log origin=infty,
	log ticks with fixed point
]
\addplot coordinates {(ANN,1.24) (MFE+A,4.26) (MFE+,2.32)};
\addplot coordinates {(ANN,0.936851960117688) (MFE+A,0.130420791205654) (MFE+,0.462551595799396)};
\addplot coordinates {(ANN,5.76) (MFE+A,2421.52) (MFE+,211.9)};
\end{semilogyaxis}
\end{tikzpicture}
\caption{HAIL7}
\end{subfigure}
\hspace{0.5cm}
%
\begin{subfigure}{0.45\textwidth}
\pgfplotsset{width=7.2cm}
\begin{tikzpicture}
\begin{semilogyaxis} [
	symbolic x coords={ANN,MFE+A,MFE+},
	xtick distance=1,
	enlargelimits=0.15,
	y tick label style={/pgf/number format/fixed},
	ybar,
	bar width=9pt,
	legend style={at={(0.5,-0.15)},	anchor=north,legend columns=-1},
	log origin=infty,
	log ticks with fixed point
]

\addplot coordinates {(ANN,1.58) (MFE+A,6.04) (MFE+,3.28)};
\addplot coordinates {(ANN,0.853333217275101) (MFE+A,0.0781376208419793) (MFE+,0.351412365428715)};
\addplot coordinates {(ANN,35.16) (MFE+A,73879.14) (MFE+,4319)};
\end{semilogyaxis}
\end{tikzpicture}
\caption{HAIL10}
\end{subfigure}
\caption{Error measures for each of the six benchmark networks. For each algorithm (ANN, MFE+A, MFE+) three error measures
were calculated: hamming distance, probability ratio, and probability rank. Note that for hamming and rank lower is better and for ratio a number closer to 1 is better.}
\definecolor{myblue}{rgb}{0.7, 0.7, 1}
\definecolor{myred}{rgb}{1, 0.7, 0.7}
\definecolor{mybrown}{rgb}{0.93, 0.85, 0.78} 
\begin{tikzpicture}
\begin{axis}[
    hide axis,
    xmin=0, xmax=2, ymin=0, ymax=2,
    legend style={at={(0.5,1.1)}, anchor=south, legend columns=-1}
]
\addlegendimage{area legend,fill=myblue}
\addlegendentry{Hamming}
\addlegendimage{area legend,fill=myred}
\addlegendentry{Ratio}
\addlegendimage{area legend,fill=mybrown}
\addlegendentry{Rank}
\end{axis}
\end{tikzpicture}
\label{ErrorMetrics_MFE+A}
\vspace{1mm}
\end{figure}
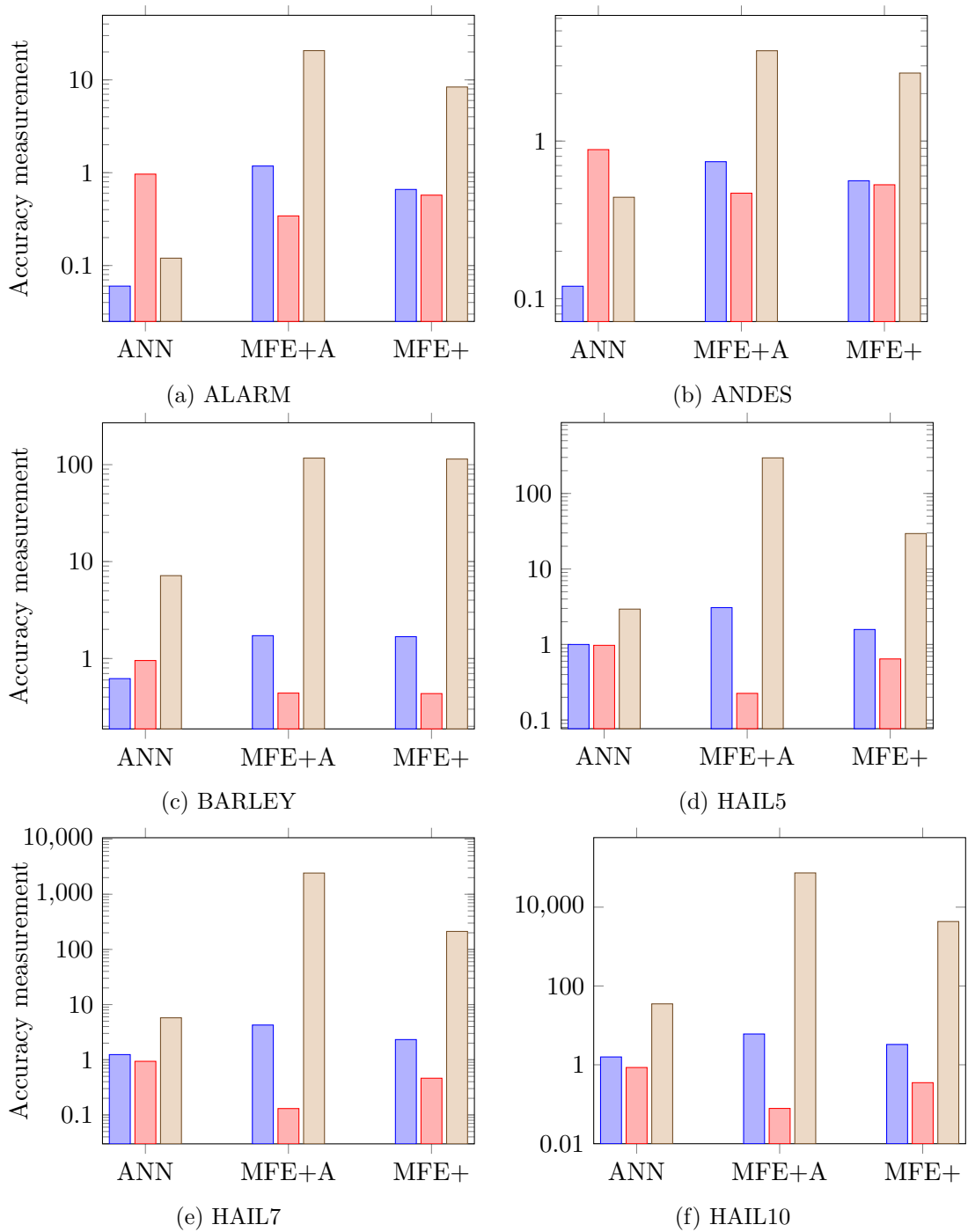

\subsection{Discussion}
\label{Discussion}

Based on the above results, we conclude that background knowledge {\em may} help, and perhaps {\em should} help in theory, but it does not really show in the current study using the LibDAI library. This is partially due to the inefficient MAP computation used as sub-routine in MFE; in addition, marginalization might be less costly for variables with small domains than sampling might be due to various overhead. When comparing actual reported running times (for {\sc Alarm} and {\sc Hailfinder}) we see that \cite{Yuan04} is about $2.5$ times faster than our results, despite running on an older computer with less memory. Perhaps the tight integration with the SMILE inference engine in this implementation can account for this. 

\section{Conclusion}
\label{Conclusion}

When reflecting back on the original research questions, it is clear that when used as approximation algorithm, MFE is unlikely to beat the state-of-the-art approximation as already for a single sample and for very few relevant variables the assessment of relevance is very costly save in the easiest situations. However, background knowledge (either precomputed or by estimation of experts, inasmuch as this is possible) may still be useful, although our results show little improvement over computing MAP exactly; however, we argued that this may be partially due to the inefficient MAP computation using the LibDAI library. We see that there is some variation on the relevant variables given the specific evidence, but there is also quite a lot of overlap so many variables may not be that relevant anyway for designated sets of hypothesis and evidence variables.

In their study comparing {\sc Annealed MAP} with {\sc P-LOC} and {\sc P-SYS}, \cite{Yuan04} uses 20 MAP variables, which favours local search approaches over approaches (such as MFE) that work on the entire hypothesis set simultaneously; however, MAP is intractable in general even for a singleton binary MAP variable, implying that there are hard instances with a limited number of hypothesis variables. It would be interesting to compare the approaches on such instances.

In regards to \textbf{RQ 3}, seems that the initial hypothesis that a combined ANN and MFE+ algorithm would result in reduced running time but poor accuracy has been proven by the results presented here. However, it was not clear exactly how much accuracy would be traded for reduced running time \textit{a priori}. After analyzing the results it appears that the drop in accuracy is so large when using MFE+A when compared to vanilla ANN or MFE+ that it possibly renders the algorithm virtually un-useable in practice, except for perhaps on very large networks such as {\sc Hailfinder10} in situations where large errors are acceptable.

For future work we obviously would like to either patch the LibDAI library with a `true' Junction-tree based MAP algorithm, or use another library that has this feature, to be able to offer MFE a fairer comparison with other approximation algorithms. We also aim to implement other approximation algorithms, such as {\sc Anytime Approximate MAP} and {\sc P-Loc} and compare MFE on other - and larger - networks. Finally, the MFE algorithm should be able to circumvent incompatible evidence due to deterministic variables so that the heuristic also works on such networks without patching.

\section*{Acknowledgements}
This paper is an extended version of \cite{Kwisthout22}. We are grateful for many relevant comments and suggestions by both the reviewers of that version as well as conference participants.

\vskip 0.2in
\bibliography{references}

\end{document}